\documentclass{article}
\usepackage{ijcai22}

\usepackage{times}

\usepackage{soul}
\usepackage{url}
\usepackage[hidelinks]{hyperref}
\usepackage[utf8]{inputenc}
\usepackage[small]{caption}
\usepackage{graphicx}
\usepackage{svg}
\usepackage{amsmath}
\usepackage{booktabs}
\usepackage{mathrsfs}
\usepackage{amsfonts}
\usepackage{algorithmic}
\usepackage{algorithm}
\usepackage{multirow}
\usepackage{subfig}
\usepackage{subfloat}
\usepackage{color}

\title{DearFSAC: An Approach to Optimizing Unreliable Federated Learning \\ via Deep Reinforcement Learning}

\author{
Chenghao Huang$^1$\and
Weilong Chen$^1$\footnote{Contact Author}\and
Yuxi Chen$^1$\and
Shunji Yang$^1$\And
Yanru Zhang$^{1,2}$\\
\affiliations
$^1$University of Electronic Science and Technology of China\\
$^2$Shenzhen Institute for Advanced Study, UESTC\\
\emails
zydhjh4593@gmail.com,
\{chenweilong1995, yuxi.ch\}@std.uestc.edu.cn, \\
shunjiy@163.com, yanruzhang@uestc.edu.cn
}

\begin{document}

\maketitle

\begin{abstract}
In federated learning (FL), model aggregation has been widely adopted for data privacy. In recent years, assigning different weights to local models has been used to alleviate the FL performance degradation caused by differences between local datasets. However, when various defects make the FL process unreliable, most existing FL approaches expose weak robustness. In this paper, we propose the DEfect-AwaRe federated soft actor-critic (DearFSAC) to dynamically assign weights to local models to improve the robustness of FL. The deep reinforcement learning algorithm soft actor-critic is adopted for near-optimal performance and stable convergence. Besides, an auto-encoder is trained to output low-dimensional embedding vectors that are further utilized to evaluate model quality. In the experiments, DearFSAC outperforms three existing approaches on four datasets for both independent and identically distributed (IID) and non-IID settings under defective scenarios.
\end{abstract}

\section{Introduction}
With the development of mobile devices, huge quantities and diverse types of data have been generated, which promote the utilization of machine learning technologies. However, when aggregated to central model training, data with sensitive privacy can lead to serious privacy leakage. To address the privacy challenges, federated learning (FL) has been proposed to aggregate local model parameters, which are only trained on local raw data, into a global model to improve performance. In FL, mobile devices are set as clients and upload their own models to the server. As a decentralized paradigm, FL significantly reduces the risks of privacy leakage by allowing clients to access only their own raw data \cite{zhu2018blockchain}.

Although FL realizes both efficient data utilization and data privacy protection in the application of mobile networks, it is fragile when various defects affect the global model during a FL process \cite{shayan2018biscotti}, such as malicious updates, poisoning attacks \cite{fung2018mitigating}, low-quality data, and unstable network environments. Unfortunately, conventional approaches pay little attention to most defects \cite{fung2018mitigating}. Therefore, an efficient approach to alleviating performance degradation caused by defective local models is strongly needed for FL. Existing researches on blockchain-based FL have defined the concept of reputation, which manifests the reliability of each local model \cite{kang2019incentive} \cite{8994206}. Similarly, we evaluate the model quality to measure how trustworthy a local model is. After learning about the quality of each local model, we are motivated to design a deep neural network (DNN) to assign optimal weights to local models, so that the global model can maintain a considerable performance no matter if there exist defects or not.

In this paper, we propose DEfect-AwaRe federated soft actor-critic (DearFSAC), a novel FL approach based on deep reinforcement learning (DRL) to guarantee a good performance of FL process by dynamically assigning optimal weights among defective local models through model quality evaluation. Since DRL algorithms often fall into local optimum, we adopt soft actor-critic (SAC) \cite{haarnoja2018soft} to find near-optimal solutions for more stable performance. Besides, as unbalanced data distribution in the buffer may deteriorate the training process of the DRL model, prioritized experience replay (PER) \cite{schaul2015prioritized} and emphasizing recent experience (ERE) \cite{wang2019boosting} are employed, which are two popular importance sampling techniques. Furthermore, as local raw data is not accessible for the server, high-dimensional model parameters trained on local raw data become the only alternative uploaded and fed into the DRL model. To avoid the curse of dimensionality, we design an embedding network using the auto-encoder framework \cite{song2013auto} to generate low-dimensional vectors containing model-quality features.

In summary, the main contributions of this paper are as follows:
\begin{enumerate}
    \item As far as we know, we are the first to propose the approach that dynamically assigns weights to local models under defective scenarios based on DRL.
    \item We design an auto-encoder based on network embedding techniques \cite{cui2018survey} to evaluate the quality of local models. This module also accelerates the DRL training process.
    \item The experimental results show that DearFSAC outperforms existing approaches and achieves considerable performance while encountering defects.
\end{enumerate}

\section{Preliminaries}
\subsection{Federated Learning}
Suppose we have one server and $N$ clients whose data $\xi_i$ is sampled from the local raw dataset $D_{i}$. The model parameters of the $i$th client and the server are denoted as $w^i_{t} \in \mathbb{R}^{d}$ and $w^g_{t} \in \mathbb{R}^{d}$ at round $t$ respectively, where $d$ is the total parameter number of one model. Then, the objective of clients is converted into an empirical risk minimization \cite{yang2019federated} as follows:
\begin{equation}
     \min_{w^{g} \in \mathbb{R}^{d}} \Big[F(w^g) = \frac{1}{N} \sum^{N}_{i=1}f_{i}(w^{g})\Big],
\label{fl goal}
\end{equation}
\begin{equation}
    f_{i}(w^{g})=\mathbb{E}_{\xi_i \sim D_{i}}[f(w^{g},\xi_i)],
\end{equation}
where $w^{g}$ is downloaded from the server by clients, and $f_{i}(w_{g})$ represents the loss of $w^g$ on the local data sampled from $D_i$. For the server, the objective is to find the optimal global model parameters:
\begin{equation}
    {w^g}^* = \arg \min_{w^{g} \in \mathbb{R}^{d}} {F(w^g)}.
\end{equation}

It is worth mentioning that federated averaging (FedAvg) \cite{mcmahan2017communication}, one of the most common used FL algorithms, simply averages all model parameters at each round. We compare FedAvg with our approach in Section \ref{experiment}.

\subsection{Deep Reinforcement Learning}
In DRL, an agent, which is usually in a DNN form, interacts with the environment by carrying out actions and obtaining rewards. The whole process can be modelled as a Markov decision process (MDP) \cite{sutton2018reinforcement}, defined by $< \mathcal{S}, \mathcal{A}, \mathcal{P}, \mathcal{R}, \gamma >$, in which $\mathcal{S}$ denotes a set of states and $\mathcal{A}$ denotes a set of actions. $\mathcal{P}: \mathcal{S} \times \mathcal{A} \times \mathcal{S} \to [0,1]$ is the state transition function used to compute the probability $p(s_{t+1}|s_t,a_t)$ of the next state $s_{t+1}$ given current action $a_t \in \mathcal{A}$ and current state $s_t \in \mathcal{S}$. The reward $r_t$ at time step $t$ is computed by the reward function $\mathcal{R}: \mathcal{S} \times \mathcal{A} \times \mathcal{S} \to \mathbb{R}$ and future rewards are discounted by the factor $\gamma \in [0,1]$. 

At each time step $t$, the agent observes the state $s_{t}$, and then interacts with environment by carrying out an action $a_{t}$ sampled from the policy $\pi(a_t|s_t):\mathcal{A} \times \mathcal{S} \to [0,1]$ which is a distribution of $a_t$ given $s_t$. After that, the agent obtains a reward $r_t$ and observes the next state $s_{t+1}$. The goal is to find an optimal policy $\pi^*$ which maximizes the cumulative return: $G=\sum^{\infty}_{t=0}\gamma^{t}r_{t}$.

\begin{figure}[ht]
\centering
\includegraphics[width=.45\textwidth]{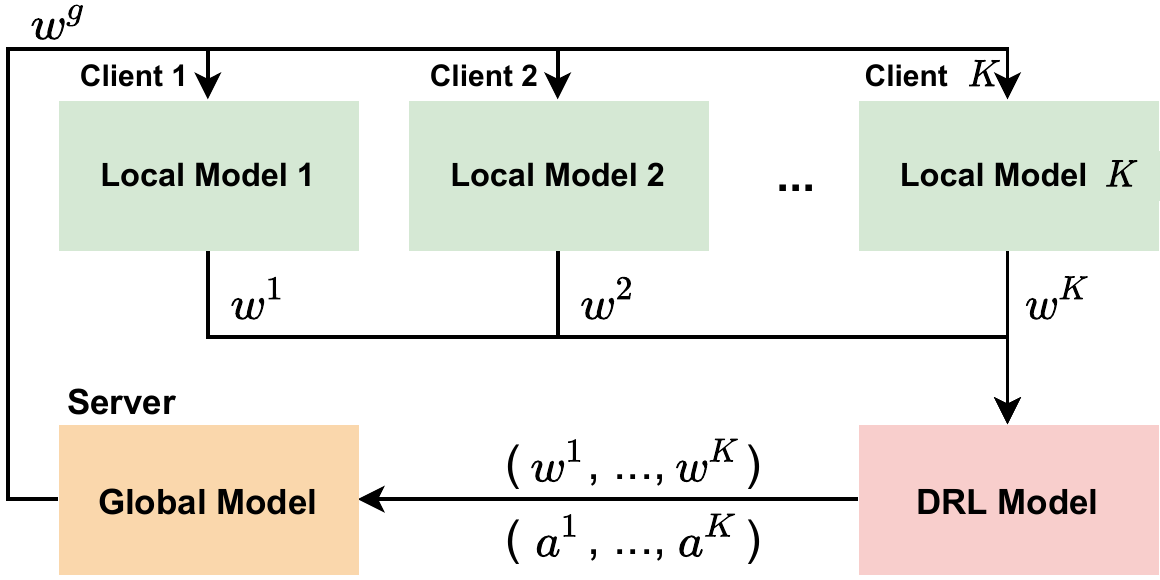}
\caption{The structure of FL combined with DRL. First, model parameters $(w^1,...,w^K)$ are randomly selected and uploaded, where $K \in \{1,..,N\}$ is usually set as $10\%$ of $N$. After model aggregation using the action from the DRL model, the server obtains the global model parameters $w^g$. Then clients download $w^g$ to update their own models.}
\label{DRL_FL}
\end{figure}

In this paper, SAC algorithm is adopted, which is an algorithm to optimize $\pi$ using actor-critic algorithm \cite{konda2000actor} and entropy regularization \cite{nachum2017bridging}. Through the DRL model, the server in FL assigns optimal weights to clients. The structure of FL combined with DRL is illustrated in Fig. \ref{DRL_FL}.

\section{Methodology}

In this section, we discuss the details of DearFSAC, a DRL-based approach to assigning optimal weights to defective local models in FL. In Section \ref{overall}, we describe the entire process of our approach. In Section \ref{QEEN-sec}, we design the quality evaluation embedding network (QEEN) for dimension reduction and model quality evaluation. In Section \ref{SAC-sec}, we adopt SAC to optimize $\pi$, which gets more stable convergence and more sufficient exploration than other actor-critic algorithms.

\subsection{Overall Architecture of DearFSAC}\label{overall}
\begin{figure*}[ht]
\centering
\includegraphics[width=.95\textwidth]{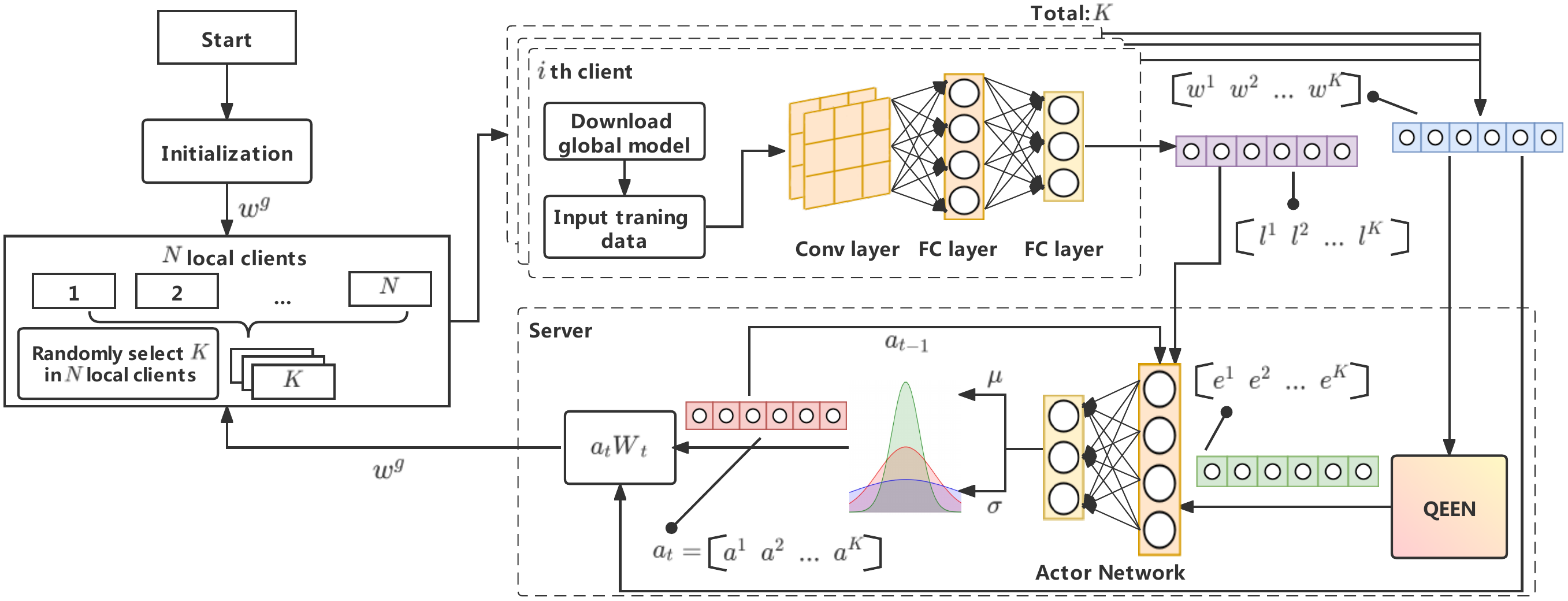}
\caption{The overall architecture of DearFSAC, in which QEEN is introduced in Fig. 3.}
\label{DearFSAC}
\end{figure*}

The overall architecture of DearFSAC is shown in Fig. \ref{DearFSAC}. At the first round, the global model parameters and the DRL action are randomly initialized. Then all clients train their own models locally and $K$ of them are randomly selected to upload model parameters $(w^1,...,w^K)$ and local training loss $(l^1,...,l^K)$. After receiving uploaded information, the server feeds the local model parameters into QEEN and gets the embedding vectors $(e^1,...,e^K)$. Next, the embedding vectors, local losses, and the last action $a_{t-1}$ are concatenated and fed into the actor network of the DRL model to get the current action $a_t$. Finally, by using $a_t$, the server aggregates local model parameters to the global model parameters $w^g$ and shares $w^g$ with all clients. The whole process loops until convergence.

\subsection{Dimension Reduction and Quality Evaluation}\label{QEEN-sec}
Based on an auto-encoder structure, QEEN is designed for both dimension reduction and quality evaluation. At round $t$, for training efficiency, we upload all local model parameters $W_{t}=(w^1_{t},...,w^N_{t})$ to the server and add several types of defects into half of them. Then we design loss $l_1$ for the embedding of $w^i_t$ and loss $l_2$ \cite{hastie2009overview} for quality evaluation. The auto-encoder on the server receives model parameters as training data and performs training using both $l_1$ and $l_2$.

We feed each $w^i_{t} \in W_{t}$ into the encoder $f_{\text{Enc}}$ composed of two fully connected (FC) layers and get the embedding vector $e^i_{t}$ of the $i$th model:
\begin{equation}
    e^i_{t} = f_{\text{Enc}}(w^i_{t}).
\end{equation}
After obtaining all embedding vectors, we put $E_{t}=(e^1_{t},...,e^N_{t})$ into the decoder to produce a decoded representation $W'_t$ which approximates $W_t$. Different from conventional ways of auto-encoder, we adopt network embedding \cite{wang2016structural} and design the decoder into parallel FC layers $f_{\text{Dec}}=(f^{1}_{\text{Dec}},...,f^{\bar{K}}_{\text{Dec}})$, where $\bar{K}$ is the number of layers of the original model, $f^{k}_{\text{Dec}}$ is the $k$th parallel FC layer corresponding to the $k$th layer of the original model structure \cite{zinkevich2010parallelized}, where $k \in \{1,..,\bar{K}\}$. Next, for the $i$th model, the embedding vector $e^i_{t}$ is fed into the $k$th parallel layer to get decoded layer parameters of the original model:
\begin{equation}
    w^{\{i,k\}}_{t} = f^{k}_{\text{Dec}}(e^i_{t}),
\end{equation}
and concatenate each $w^{\{i,k\}}_{t}$ layer by layer to obtain the entire decoded model parameters:
\begin{equation}
    {w^i_{t}}' = \textbf{Concat}(w^{\{i,1\}}_{t},...,w^{\{i,k\}}_{t}).
\end{equation}
After getting $W'_{t}=({w^1_{t}}',...,{w^K_{t}}')$, we use mean square error (MSE) loss function to compute $l_1$:
\begin{equation}
    l_1 = \text{MSELoss}(W'_{t}, W_t) = \frac{1}{K} \sum^{K}_{i=1}({w^i_t}'-w^i_t)^2.
\end{equation}

As multiple defects have different impact on local models, we define defect marks as the ground truth, denoted as $\mathcal{N}_t=\{n^1_{t},...,n^N_{t}\}$, where $n^1_t$ is the degree of defect. Next, we compare defect marks with quality evaluation marks $\mathcal{N}'_t=\{{n^1_{t}}',...,{n^N_{t}}'\}$. We feed $e^i_{t}$ into the quality evaluation module $f_{\text{QE}}$ composed of two FC layers to get ${n^i_t}'$ to predict the quality of the $i$th model:
\begin{equation}
    {n^i_t}' = f_{\text{QE}}(e^i_{t}).
\end{equation}
Then we compute $l_2$:
\begin{equation}
    l_2 = \text{MSELoss}(\mathcal{N}'_{t}, \mathcal{N}_t) = \frac{1}{K} \sum^{K}_{i=1}({n^i_t}'-n^i_t)^2.
\end{equation}

\begin{figure}[ht]
\centering
\includegraphics[width=.45\textwidth]{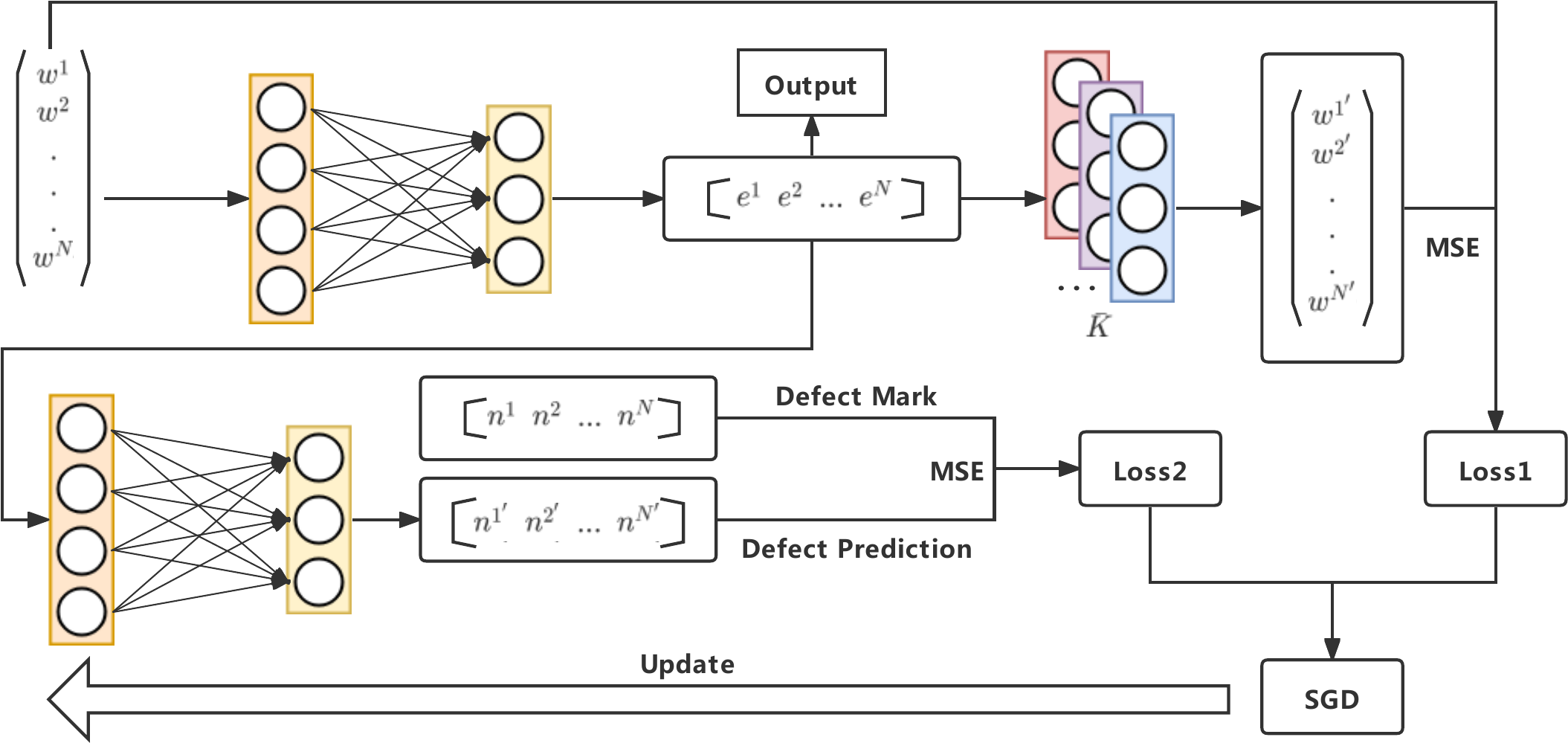}
\caption{The structure of QEEN. Model parameters $(w^1,...,w^N)$ are fed into the encoder. Then the embedding vectors $(e^1,...,e^N)$ are outputted and used to update QEEN.}
\label{QEEN}
\end{figure}

Finally, we set different weights $\lambda^{\text{QEEN}}_{1}$ and $\lambda^{\text{QEEN}}_{2}$ for two kinds of loss, generally $0.5$ and $0.5$ respectively, to update the QEEN parameter $\theta_{\text{QEEN}}$ using joint gradient descent \cite{tanaka2018joint} as follows:
\begin{equation}
    \theta_{\text{QEEN}} = \theta_{\text{QEEN}} - \lambda^{\text{QEEN}}_{1}\bigtriangledown_{\theta_{\text{QEEN}}} l_{1}-\lambda^{\text{QEEN}}_{2}\bigtriangledown_{\theta_{\text{QEEN}}} l_{2}.
\end{equation}
The entire training process of QEEN is illustrated in Fig. \ref{QEEN}.

\subsection{DRL for Optimal Weight Assignment}\label{SAC-sec}
\subsubsection{MDP Modelling:}
To guarantee the communication efficiency and fast convergence, $K$ clients are randomly selected among the $N$ clients at each round $t$ and upload models to the server. After receiving various information as the current state, the DRL model outputs an action containing weights of all selected models. The details and explanations of $\mathcal{S}$, $\mathcal{A}$, and $\mathcal{R}$ are defined as follows:

\textbf{State} $\mathcal{S}$: At round $t$, the state $s_{t}$ can be denoted as a vector $(e^g_{t}, e^1_{t},..., e^K_{t}, l^1_{t}, ..., l^K_{t}, \mathbf{a}_{t-1})$, where $e^i_{t}$ denotes the embedding vector of $i$th client's model parameters,  $e^g_{t}$ denotes the embedding vector of the server's model parameters, $l^i_{t}$ denotes the local training loss of $i$th local model, and $a_{t-1}$ denotes the action at the last round.

\textbf{Action} $\mathcal{A}$: The action, denoted as $\mathbf{a}_{t}=\{a^1_{t}, a^{2}_{t},..., a^K_{t}\}$, is a weight vector calculated by the DRL agent for randomly selected subset of $K$ model parameters at round $t$. All the weights in $\mathbf{a}_{t}$ are within $[0,1]$ and satisfy the constraint $\sum^{K}_{i=1}a^i_{t} = 1$. After obtaining the weight vectors, the server aggregates local model parameters to the global model as follows:
\begin{equation}
\begin{aligned}
w^g_{t} = \mathbf{a}_{t} W_{t} = \sum^{K}_{i=1} a^i_{t} w^i_{t},
\end{aligned}
\end{equation}
where $W_{t} \in \mathbb{R}^{Kd}$ is a set of all selected local models. 

\textbf{Reward} $\mathcal{R}$: The goal of DRL is to maximize cumulative reward $R$ in total time steps $T$, which is equivalent to finding the local model with minimum loss shown in Eq. (\ref{fl goal}). Therefore, we design a compound reward by combining three sub-rewards with appropriate weights $\beta_{i}$, which can be formulated as:
\begin{align}   
R &= \sum^{3}_{i=1} \sum^{T}_{t=1}\gamma^{t-1} \beta_{i} r^i_{t}.\\
\label{r1} r^1_{t}&=\begin{cases}
    \kappa^{\delta_{t}-\bar{\delta}_{t}} - 1, \quad \text{if} \quad \delta_{t} < 0.5 , \\
	\kappa^{\delta_{t}-\Delta} - 1, \quad \text{else},
  \end{cases} \\
\label{r2} r^{2}_{t}&= - \frac{1}{K} \sum^{K}_{i=1} {({\bar{n}^i_t} - a^i_t)}^2, 
\end{align}

\begin{equation}
\begin{aligned}
\label{r3} r^{3}_{t} &= \frac{1}{2} \cos({w^g_{t},w^g_{t+1}}) - \frac{1}{2} \\
&= \frac{1}{2} \frac{\sum^{d}_{i=1}\dot{w}^g_{t}\dot{w}^g_{t+1}} {\sqrt{\sum^{d}_{i=1}(\dot{w}^{g}_{t})^2}\sqrt{\sum^{d}_{i=1}(\dot{w}^g_{t+1})^2}} - \frac{1}{2}.
\end{aligned}
\end{equation}

In Eq. (\ref{r1}), $r^1_t$ is defined within $(-1,0]$ to maximize global model's accuracy. The exponential term $\delta_t - \bar{\delta}_t$ and $\delta_t - \Delta$ represent the accuracy gap, where $\delta_{t}$ is the global model's accuracy on the held-out validation set at round $t$, $\Delta$ is the target accuracy which is usually set to $1$, and $\bar{\delta}$ is the accuracy of the model aggregated by FedAvg. $\kappa$ is a positive constant to ensure an exponential growth of $r^1_{t}$. As $\kappa^{\delta_{t}-\Delta}$ is in $(0,1]$, the second term, $-1$, is used as time penalty at each round $t$ to set $r^1_{t}$ to $(-1,0]$ for faster convergence.

Eq. (\ref{r2}) aims to provide auxiliary information for the agent to reduce exploration time. After obtaining quality prediction mark ${n^i_{t}}'$ of the $i$th local model from QEEN, we denote normalized ${n^i_{t}}'$ as ${\bar{n}^i_{t}} \in [0,1]$ to calculate the MSE loss of ${\bar{n}^i_{t}}$ and $a^i_t$. Similarly, for time penalty, Eq. (\ref{r2}) is set to be negative.

Eq. (\ref{r3}) stabilizes the agent actions by comparing $w^g_{t}$ and $w^g_{t+1}$ using cosine similarity \cite{dehak2010cosine}. Specifically, we compute $r^{3}_{t}$ using parameters of two models, which is denoted as $\dot{w}$. As $\cos({w^g_{t},w^g_{t+1}})$ is in $[-1,1]$, we use $\frac{1}{2}$ to normalize it in $[-1,0]$.

\subsubsection{Adopting SAC to Solve MDP:}
First, $K$ locally trained models are randomly selected to upload the parameters $W_{t}=(w^1_t,..., w^K_t)$ and local loss $L_t=(l^1_{t}, ..., l^K_{t})$ to the server. Through QEEN, we can obtain embedding vectors $E_{t} = (e^i_{t},..., e^K_{t})$ as part of the current state $s_{t}$. By feeding $s_{t} = (e^g_{t}, e^1_{t},..., e^K_{t}, l^1_{t}, ..., l^K_{t}, a_{t-1})$ into the actor network, we obtain the current action $a_t \sim \pi_{\phi}(s_t)$. After model aggregation, we get reward $r_t$ and the next state $s'$. Empirically, we set $(\beta_1,\beta_2,\beta_3)$ as $(0.5,0.4,0.1)$. At the end of each round, the tuple $(s,a,r,s')$, which is denoted as $\tau$, is recorded in the buffer. 

For each iteration, SAC samples a batch of $\tau$ from the buffer and updates the DRL network parameters. To deal with poor sampling efficiency and data unbalance in DRL, we adopt two techniques of replay buffer named ERE \cite{wang2019boosting} and PER \cite{schaul2015prioritized} to sample data with priority and emphasis. For the $\dot{t}$th update, we sample data uniformly from the most recent data points $c_{k}$, defined as:
\begin{equation}
    c_{k} = \max\{ |B_{\text{max}}| \cdot \eta^{\dot{t}\frac{1000}{\dot{T}}}, c_{\text{min}} \},
\end{equation}
where $\eta \in (0,1]$ represents the degree of emphasis on recent data and $|B_{max}|$ is the maximum size of buffer $B$. After obtaining an emphasizing buffer $B' \sim B$ according to $c_{k}$, the sampling probability of data point $P(i)$ in PER is computed as:
\begin{equation}\label{P-PER}
    P(i) = \frac{p^{\nu_{1}}_{i}}{\sum_{j}p^{\nu_{2}}_{j}}, \quad i,j \in B',
\end{equation}
where $\nu_{1}$ is a hyperparameter determining the affection of the priority, $\nu_{2}$ is a hyperparameter controlling the affection of $w_{i}$. $p_{i}$ in Eq. (\ref{P-PER}) is the priority value of the $i$th data point, defined as:
\begin{equation}
p_{i} = \frac{1}{2}\sum^{2}_{i=1}|R+\gamma \max_{a}Q_{i}(s',a) - Q_{i}(s,a)| + \varepsilon,
\end{equation}
where $\varepsilon$ is the bias and $Q$ is the action-value function formulated as:
\begin{equation}\label{Q}
\begin{aligned}
Q(s,a) = \mathbb{E}_{s' \sim P \atop a' \sim \pi}\bigg[R(\tau) + \gamma(Q(s',a') + \alpha \mathcal{H}\left(\pi(\cdot|s')) \right)\bigg],
\end{aligned}
\end{equation}
where $\mathcal{H}(\pi)$ is the entropy of $\pi$, formulated as:
\begin{equation}
    \mathcal{H}(\pi) = \mathbb{E}_{a \sim \pi}[-\log \pi(a)].
\end{equation}
Next, we compute the importance sampling weight $w_{i}$ of the $i$th data point as:
\begin{equation}
w_{i}=\Bigg(\frac{1}{|B|}\cdot\frac{1}{P(i)}\Bigg)^{\nu_2}.
\end{equation}

After sampling $\tau$, SAC \cite{haarnoja2018soft} updates the DRL model and aims to find $\pi^*$ to maximize both the total reward and the entropy, which leads to more stable convergence and more sufficient exploration:
\begin{equation}
\pi^* = \arg \max_{\pi} \mathbb{E}_{\tau \sim \pi}{\sum_{t=0}^{\infty} \gamma^t \bigg[ R(\tau) + \alpha \mathcal{H}\left(\pi(\cdot|s_t)\right) \bigg]},
\end{equation}
where $\alpha > 0$ is a trade-off coefficient.

\section{Experiments}\label{experiment}
In this section, we conduct various experiments to validate the performance of DearFSAC on defective local models. Specially, we compare the test accuracy of DearFSAC with different approaches on four datasets in Section \ref{performance}. Then, we try different numbers of defective models and degrees of defect to show robustness of DearFSAC in Section \ref{robust}. Besides, in Section \ref{tech}, we discuss the effectiveness of QEEN by designing ablation experiments.

\subsection{Experimental Setup}\label{setup}
\subsubsection{Datasets}
We validate the proposed DRL model on four datasets: MNIST \cite{lecun1998gradient}, CIFAR-10 \cite{krizhevsky2009learning}, KMNIST \cite{clanuwat2018deep}, and FashionMNIST \cite{xiao2017fashion}. For convenience, we call the three MNIST datasets X-MNIST. The setup is illustrated in Table \ref{FL setup}, The X-MNIST datasets contain both IID and non-IID data while the CIFAR-10 dataset contains only IID data.
\begin{table}[h!]
    \centering
    \begin{tabular}{cccc}
    \hline
    \multirow{2}*{\textbf{Parameter}}&\multicolumn{2}{c}{\textbf{X-MNIST}}& \textbf{CIFAR-10}\\ 
     & \textbf{IID} & \textbf{NonIID} & \textbf{IID}\\ \hline
    Total Clients $N$ & 100 & 100 & 100\\ \hline
    Selection Number $K$ & 10 & 10 & 10\\ \hline
    Model Size $|w|$ & 26474 & 26474 & 62006\\ \hline
    \end{tabular}
    \caption{FL setup on different datasets}
    \label{FL setup}
\end{table}

\subsubsection{Defect Types}
We define the number of defective models as $M$ and the degree of defect as $d_{\mathcal{N}}$. Then we design three types of defect:
\begin{itemize}
    \item \textbf{Data contamination:} We add standard Gaussian noise $n^G \sim \frac{1}{\sqrt{2\pi}}\exp\{-\frac{x^2}{2\sigma^2}\}$ to each pixel $p_{\text{in}}$ in an image and obtain defective pixels $p_{\text{out}} = p_{\text{in}} + n^G d_{\mathcal{N}}$.
    \item \textbf{Communication Loss:} We add standard Gaussian noise $n^G \sim \frac{1}{\sqrt{2\pi}}\exp\{-\frac{x^2}{2\sigma^2}\}$ to each parameter $w_{\text{in}}$ in the last two layers and obtain defective parameters $w_{\text{out}}=w_{\text{in}} + n^G d_{\mathcal{N}}$.
    \item \textbf{Malicious attack:} For both IID and non-IID datasets, we shuffle labels of each local training batch.
\end{itemize}

\subsubsection{Metrics}
To evaluate the performance of DearFSAC and compare it with other weight assignment approaches, we mainly identify three performance metrics as follows:
\begin{itemize}
    \item $Acc_{\text{avg}}$: The averaging accuracy on test datasets over multiple times.
    \item $T_{\Delta}$: The number of communication rounds to first achieve $\Delta$ in corresponding datasets.
    \item $G$: The cumulative reward of DRL approaches in each episode.
\end{itemize}

\subsection{Comparisons across Different Datasets}\label{performance}
In this subsection, we compare our approach with FedAvg, rule-based strategy, and supervised learning (SL) model. For rule-based strategy, it assigns weights $\frac{1}{K-M}$ to models with no defects. For SL model \cite{cui2018survey}, it consists of $2$ FC layers with $128$ and $64$ units and performs training with defect marks. We compose three types of defects at the same time to obtain the composite defect. Then we adopt it in both DRL training process and FL test. We conduct experiments on the FL training dataset for $100$ rounds, setting $M=9$ and $d_{\mathcal{N}}=0.1$.

As shown in Table \ref{dataset comparison}, we carry out 100-round FL training process for ten times and compare $Acc_{\text{avg}}$ and $T_{\Delta}$ of each approach. The results on the IID datasets show that our approach significantly outperforms the other three approaches in all four IID datasets. Furthermore, we compare our approach with FedAvg with no defect in local models and find that our approach performs almost the same as FedAvg in the defectless setting. This is because the data distribution is IID so that averaging weights is a near-optimal strategy, which exhibits that our approach converges to FedAvg in the simplest setting.

On the other hand, our approach also performs the best on non-IID datasets. As data distribution is largely different, $Acc_{\text{avg}}$ of each approach decreases obviously, especially the rule-based strategy and SL model. In non-IID KMNIST, the performance of rule-based strategy is similar to that of FedAvg. These two results show that fixed weight is not feasible in non-IID datasets. Besides, FedAvg with no defects needs more communication rounds on non-IID datasets than DearFSAC, which shows the advantage in speed of DearFSAC.

All the above results show that our approach performs the best no matter whether there exist defects in local models or not, which verify the generalization of our approach.

\begin{table*}[ht]
\centering
\begin{tabular}{ccc|cc|cc|c}
\hline
\multirow{2}{*}{Approach}          & \multicolumn{2}{c|}{MNIST}                                                 & \multicolumn{2}{c|}{KMNIST}                                                 & \multicolumn{2}{c|}{FashionMNIST}                                           & CIFAR                                \\
                                   & IID                                 & Non-IID                              & IID                                  & Non-IID                              & IID                                  & Non-IID                              & IID                                  \\ \hline
DearFSAC-nodefect                  & 97.45\%/\textbf{7} & 94.64\%/20                           & \textbf{89.03\%/39} & 76.52\%/\textbf{35} & 85.06\%/44                           & \textbf{73.98\%}/23 & \textbf{58.29\%/40} \\
\textbf{DearFSAC} & \textbf{98.06\%/7} & \textbf{95.29\%/19} & 88.69\%/40                           & \textbf{77.2\%}/36  & \textbf{85.59\%/43} & 73.47\%/\textbf{21} & 57.21\%/41                           \\
FedAvg-nodefect                    & 97.57\%/11                          & 95.07\%/20                           & 88.23\%/42                           & 75.30\%/39                           & 85.43\%/44                           & 71.69\%/26                           & 57.37\%/41                           \\
FedAvg                             & 62.76\%/-                           & 39.26\%/-                            & 42.65\%/-                            & 28.72\%/-                            & 33.61\%/-                            & 22.55\%/-                            & 28.15\%/-                            \\
Rule-based                         & 85.27\%/-                           & 69.37\%/-                            & 72.78\%/-                            & 31.29\%/-                            & 68.17\%/-                            & 26.67\%/-                            & 47.93\%/-                            \\
SL                                 & 86.20\%/-                           & 75.88\%/-                            & 78.97\%/-                            & 39.83\%/-                            & 69.51\%/-                            & 28.91\%/-                            & 51.57\%/-                            \\ \hline
\end{tabular}
\caption{$Acc_{\text{avg}}$ and $T_{\Delta}$ of DearFSAC, FedAvg, rule-based strategy and SL model on IID and non-IID datasets, where $M=9$, $d_{\mathcal{N}}=0.1$, and $\Delta$ is 95\%/90\% for the CNN on IID/non-IID MNIST, 85\%/75\% for IID/non-IID KMNIST, 85\%/70\% for IID/non-IID FashionMNIST, and 55\% for IID CIFAR-10. Also, DearFSAC and FedAvg with no defects are demonstrated in this table. Best results are in bold.}
\label{dataset comparison}
\end{table*}

\subsection{Defect Impact} \label{robust}
In this subsection, we compare the performance of the above approaches on non-IID MNIST to study the impact of different $M$ and $d_{\mathcal{N}}$.

\begin{figure*}[ht] \centering    
\subfloat {
\includegraphics[width=0.23\textwidth]{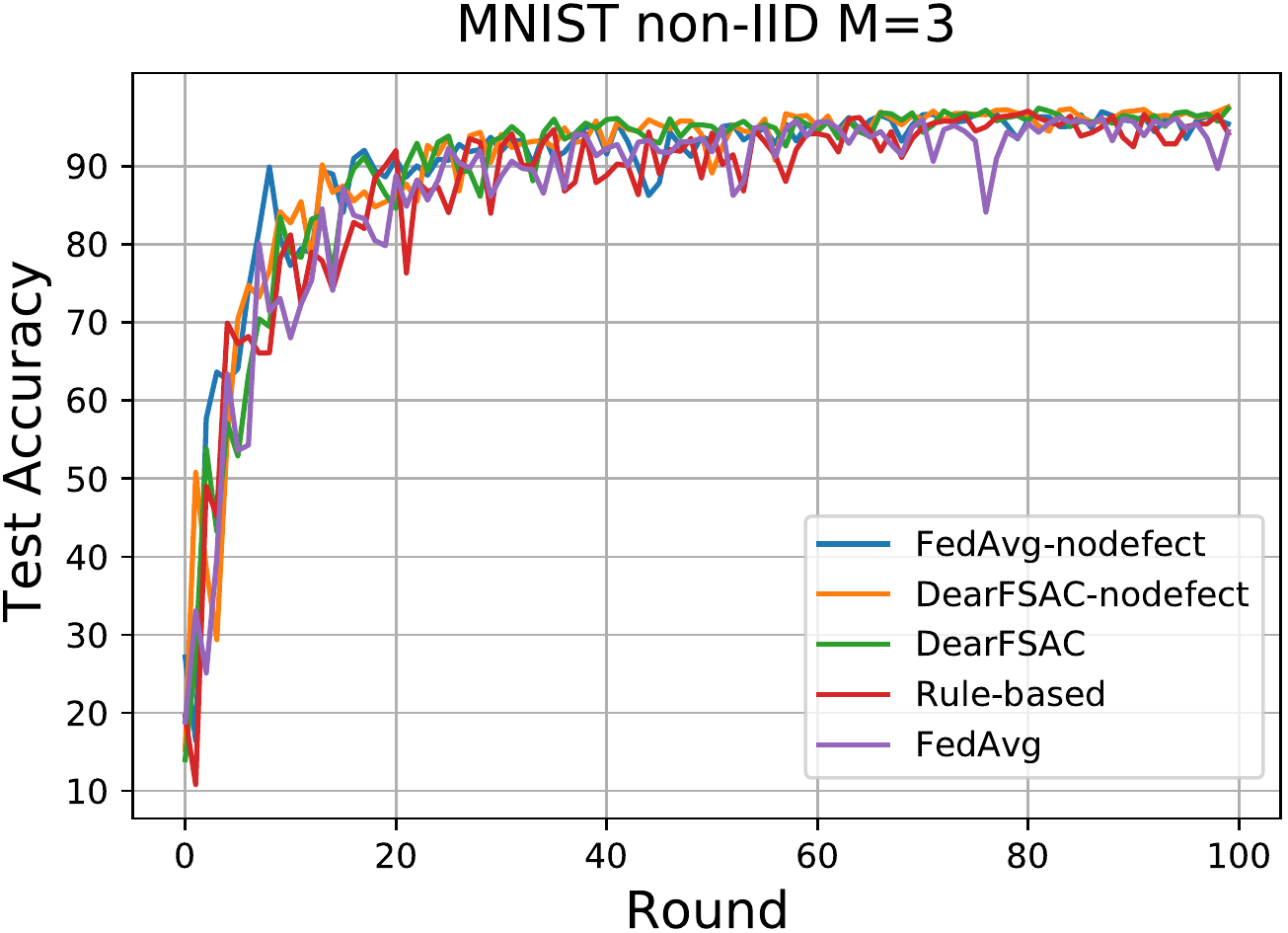}  }   
\subfloat {
\includegraphics[width=0.23\textwidth]{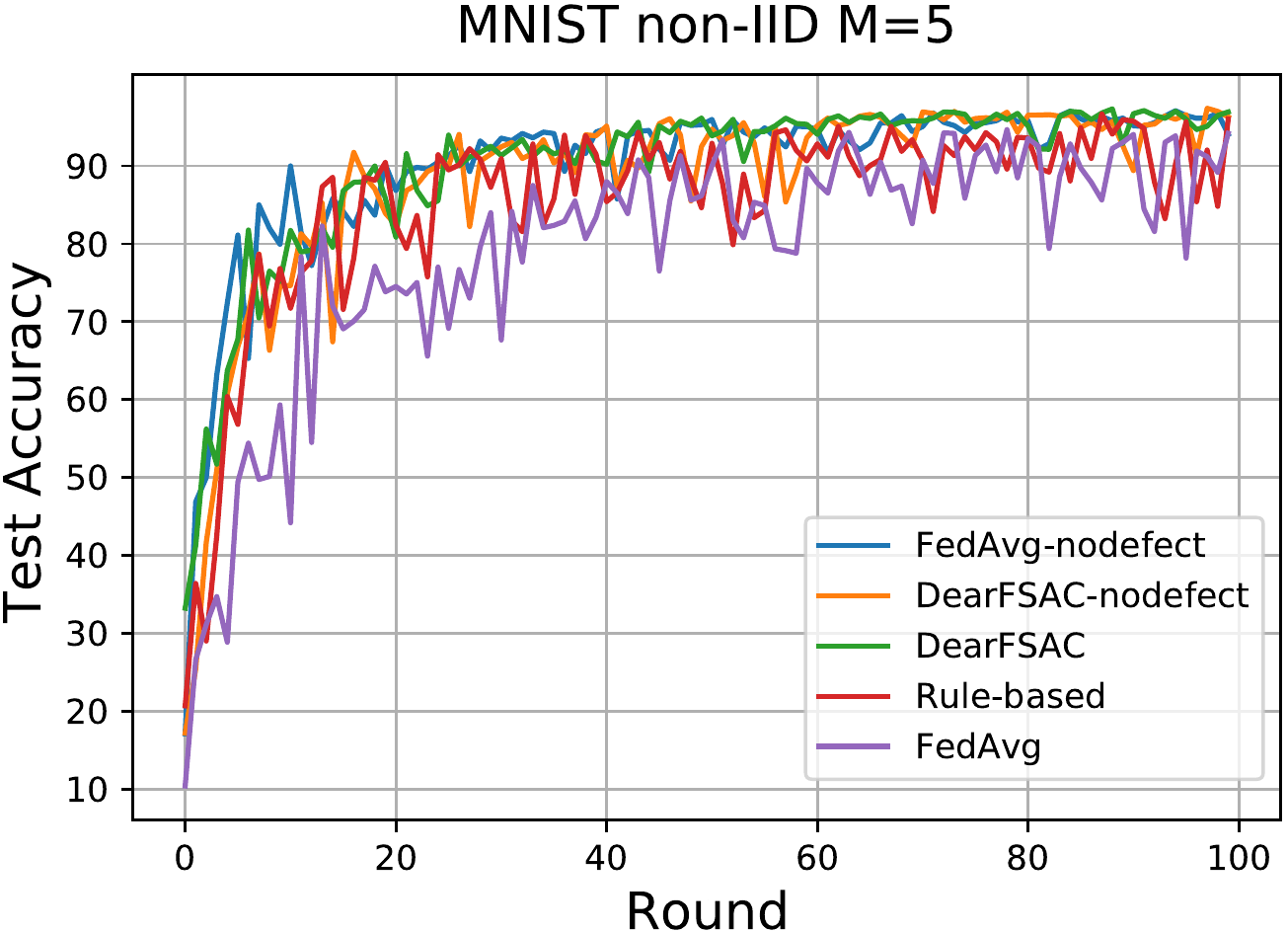}  }    
\subfloat { 
\includegraphics[width=0.23\textwidth]{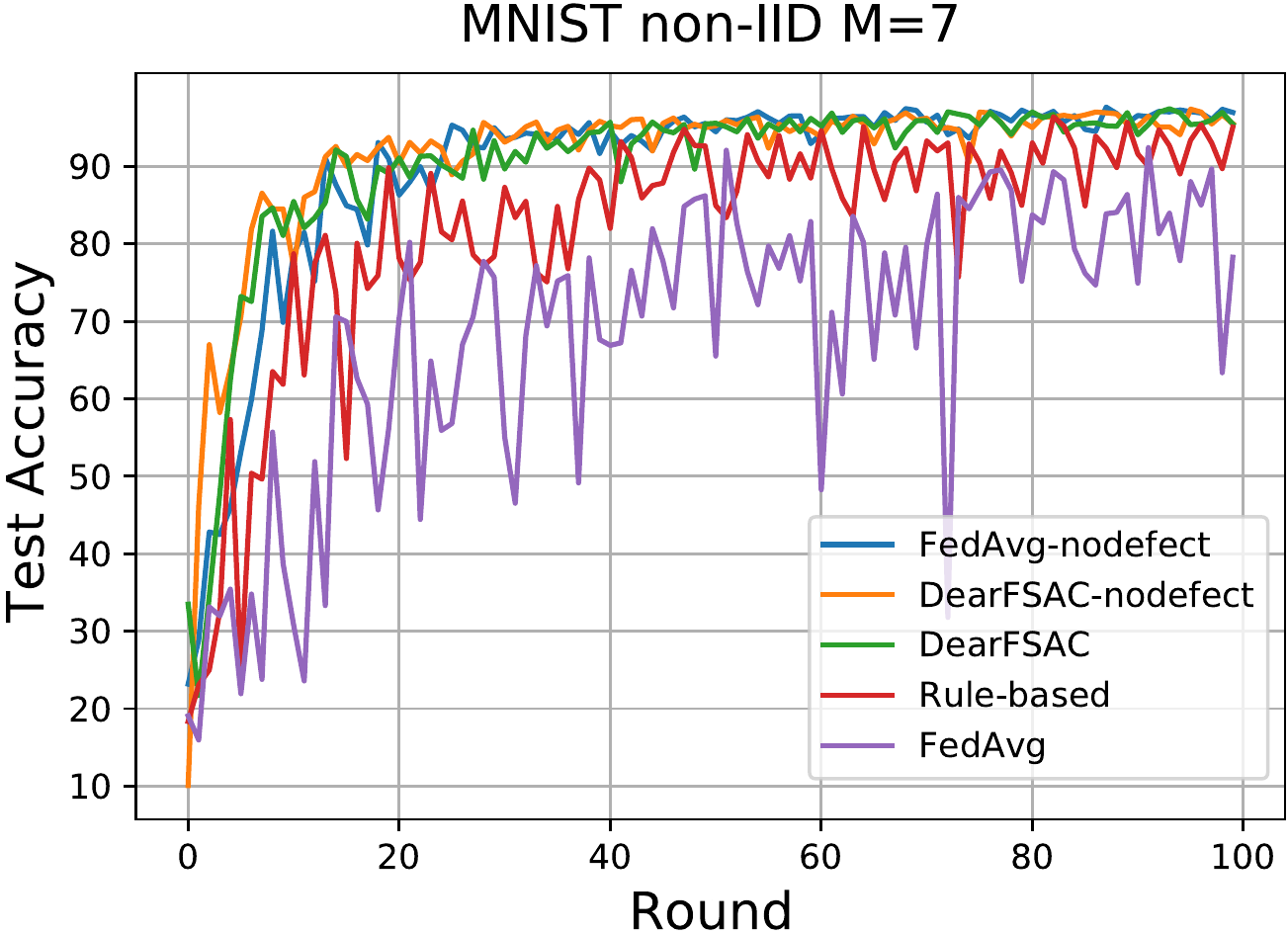}  }    
\subfloat { 
\includegraphics[width=0.23\textwidth]{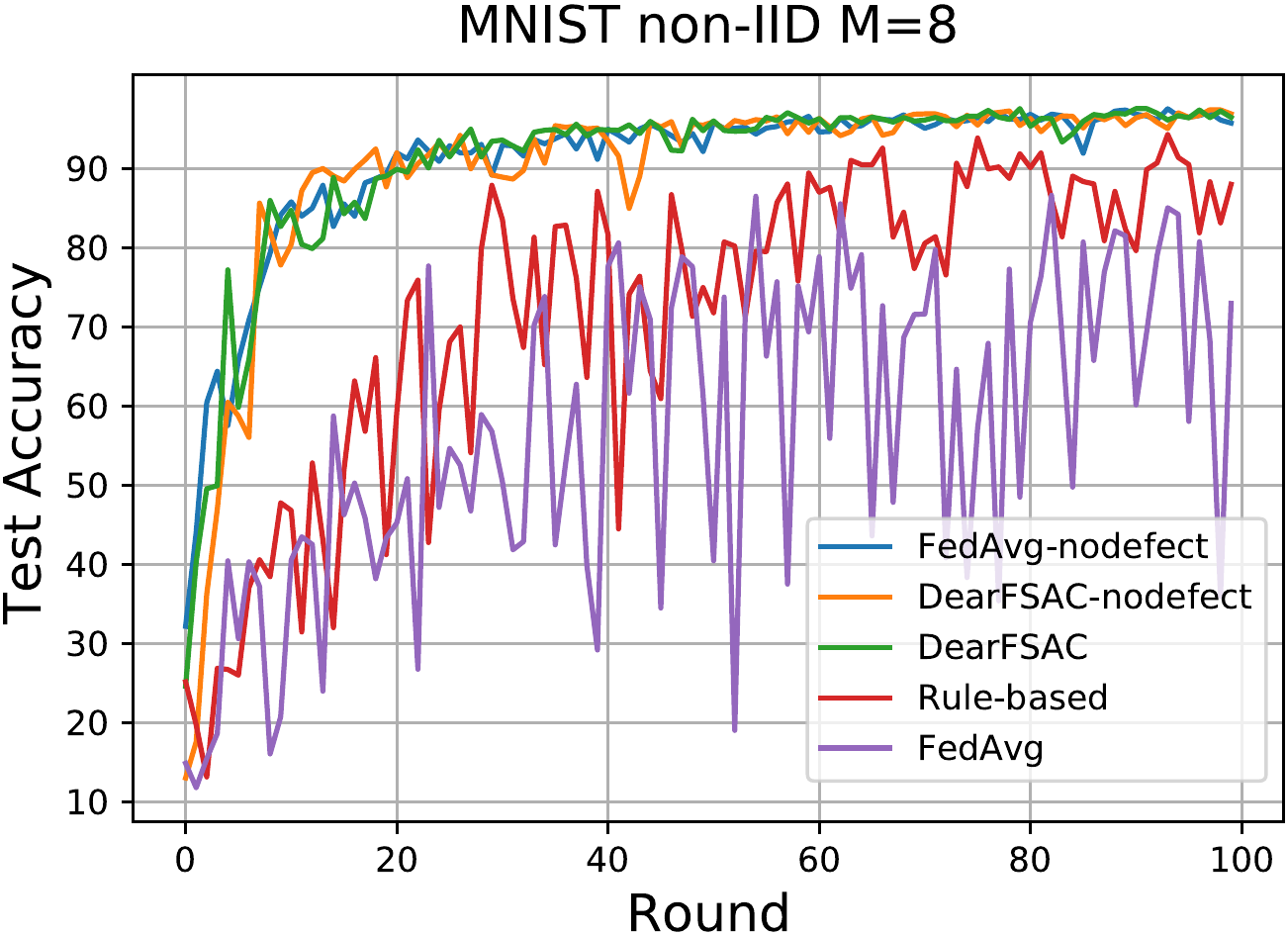}}
\caption{The accuracy of DearFSAC, FedAvg, rule-based strategy on non-IID MNIST, where $M=3,5,7,8$ and $d_{\mathcal{N}}=0.1$.}
\label{acc_amount}     
\end{figure*}

First, we change the value of $M$ to study how the numbers of defective models impact the performance. Fig. \ref{acc_amount} shows that as $M$ increases, the accuracy decreases dramatically. When $M$ is small, defects cause little impact on the global model. On the contrary, if $M$ is relatively larger, it becomes sensitive to accuracy of the global model. It also shows that FL has limited capability to resist defects. Compared with FedAvg, our approach has a more robust performance despite large $M$.

\begin{table}[]
\centering
\begin{tabular}{clcccc}
\hline
\multicolumn{2}{c}{\multirow{2}{*}{$d_{\mathcal{N}}$}} & \multicolumn{4}{c}{Approach}                \\
\multicolumn{2}{c}{}                        & \textbf{DearFSAC} & FedAvg  & Rule-based & SL      \\ \hline
\multicolumn{2}{c}{0.1}                     & 94.64\%  & 35.26\% & 69.37\%    & 75.88\% \\
\multicolumn{2}{c}{0.3}                     & 94.91\%  & 19.64\% & 68.21\%    & 71.62\% \\
\multicolumn{2}{c}{0.5}                     & 94.27\%  & 12.73\% & 68.77\%    & 63.83\% \\
\multicolumn{2}{c}{0.7}                     & 93.87\%  & 10.33\% & 70.05\%    & 53.11\% \\
\multicolumn{2}{c}{0.9}                     & 95.06\%  & 9.55\%  & 69.56\%    & 42.65\% \\ \hline
\end{tabular}
\caption{$Acc_{\text{avg}}$ of DearFSAC, FedAvg, rule-based strategy and SL model on non-IID MNIST with different $d_{\mathcal{N}}$.}
\label{degree}
\end{table}

In Table \ref{degree}, we study how the degree of the composite defect $d_{\mathcal{N}}$ affects the performance. As $d_{\mathcal{N}}$ increases, the accuracy of FedAvg decreases dramatically while our approach holds a high and stable accuracy, which indicates that the accuracy of the global model is quite sensitive to $d_{\mathcal{N}}$.

All the above experiments show that our approach is capable of adapting multiple numbers and degrees of composite defect, validating the robustness of our approach.

\subsection{Effectiveness of QEEN}\label{tech}
In this subsection, we study the effectiveness of QEEN by comparing the cumulative reward $G$ and the accuracy of DearFSAC with that of original SAC and embedding SAC, where embedding SAC adopts only an embedding network for dimension reduction. We compare the three versions of DRL model on IID and non-IID MNIST datasets. Here we set $G = \sum^{3}_{i=1} \sum^{50}_{t=1}\gamma^{t-1} \beta_{i} r^i_{t}$, where total episodes $T$ is $800$ and each episode contains $50$ rounds.
\subsubsection{Cumulative Reward} 
\begin{figure} \centering    
\subfloat { 
\includegraphics[width=0.22\textwidth]{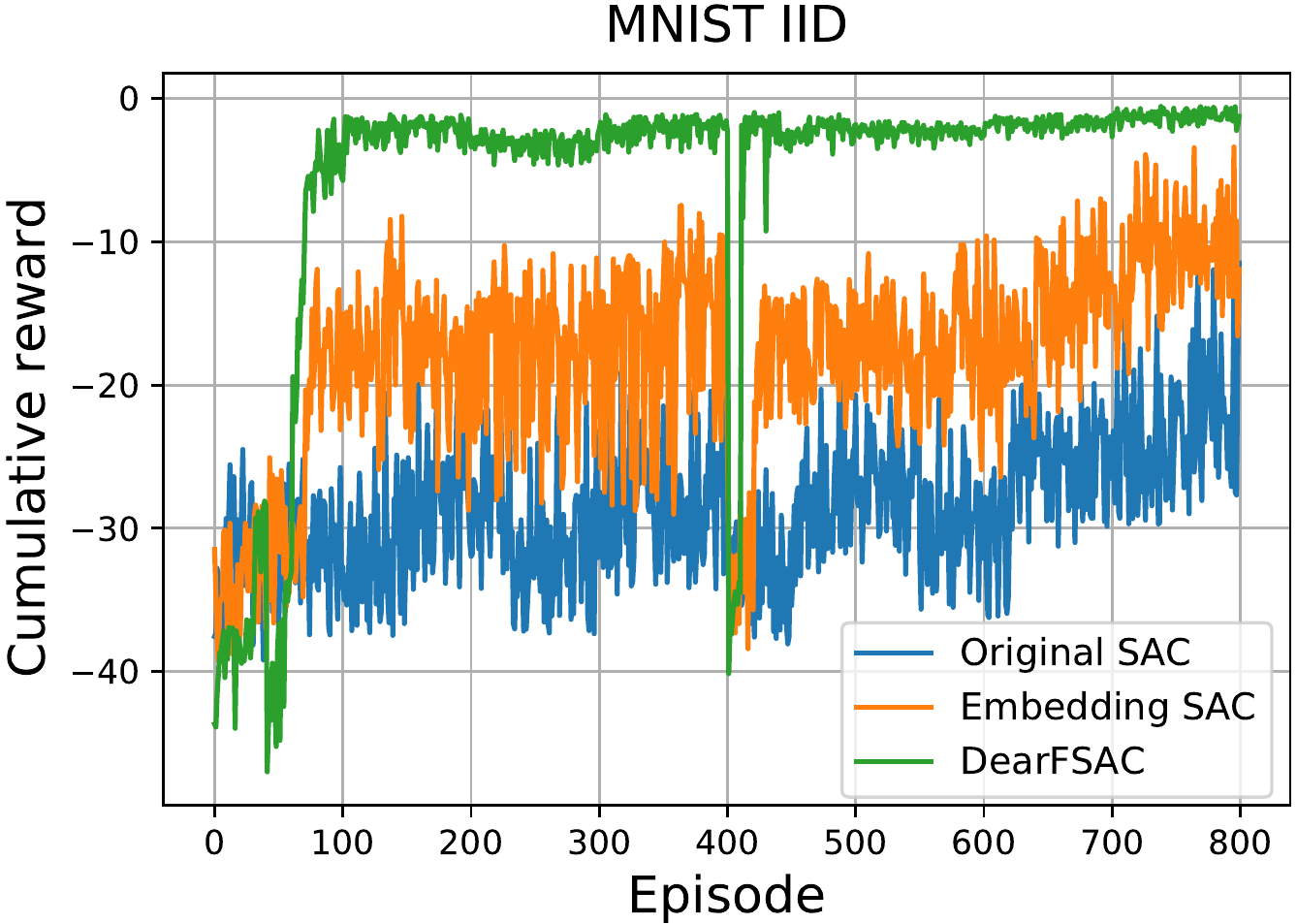}  } 
\subfloat {
\includegraphics[width=0.22\textwidth]{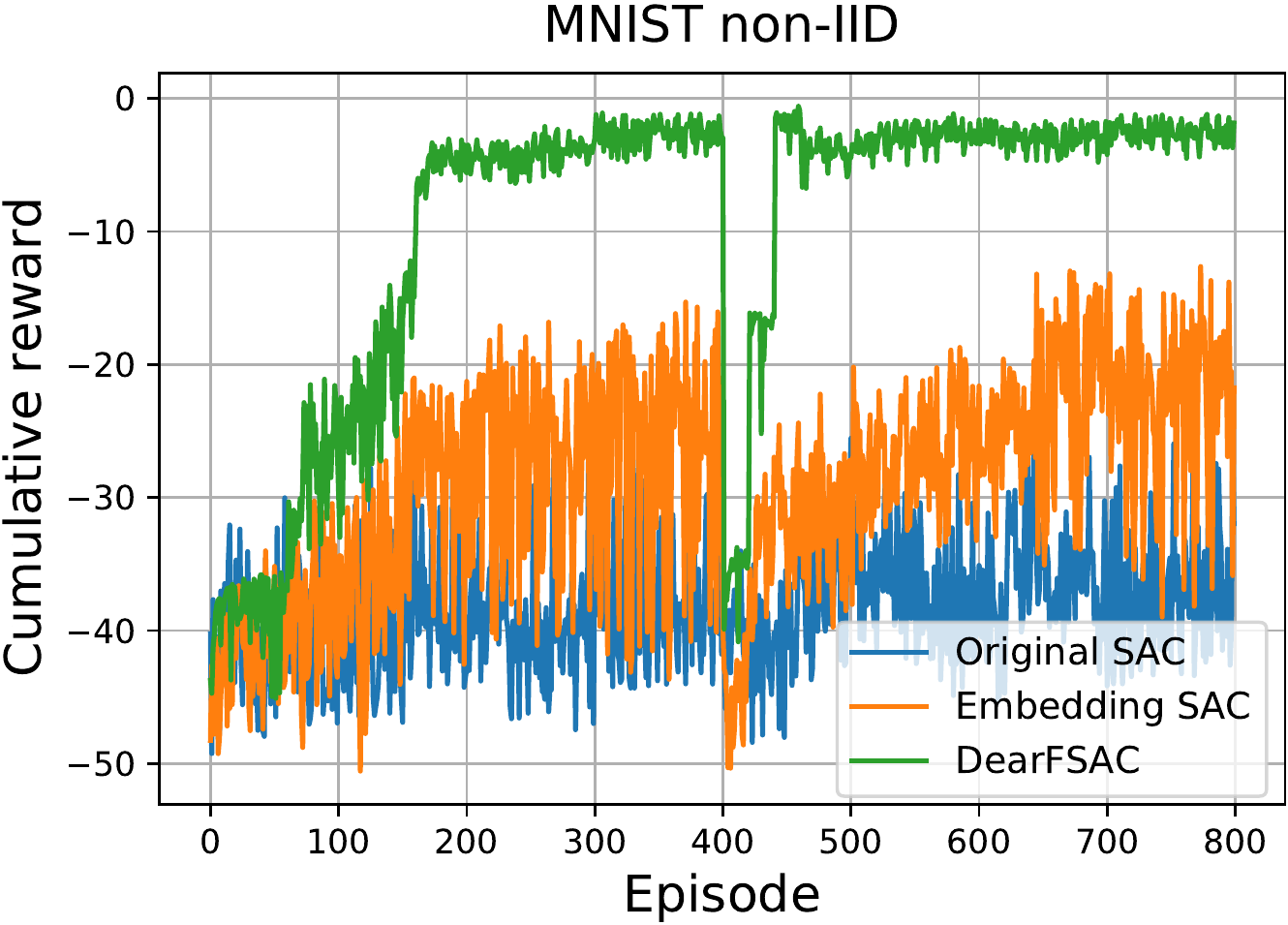}  }     
\caption{The reward of DearFSAC, embedding SAC and original SAC on IID MNIST and non-IID MNIST, where $M=5$ and $d_{\mathcal{N}}=0.1$.}   
\label{reward_sac} 
\end{figure}

In Fig. \ref{reward_sac}, for IID MNIST and non-IID MNIST, $G$ of DearFSAC increases rapidly at the beginning and gradually converges, while $G$ of embedding SAC fluctuates dramatically, which indicates that quality evaluation not only largely improves the accuracy, but also guarantees the convergence speed and stability in DearFSAC. Besides, $G$ of original SAC is the worst which means that embedding network also matters in DearFSAC for good performance.

\subsubsection{Accuracy} 
\begin{figure} \centering   
\subfloat { 
\includegraphics[width=0.21\textwidth]{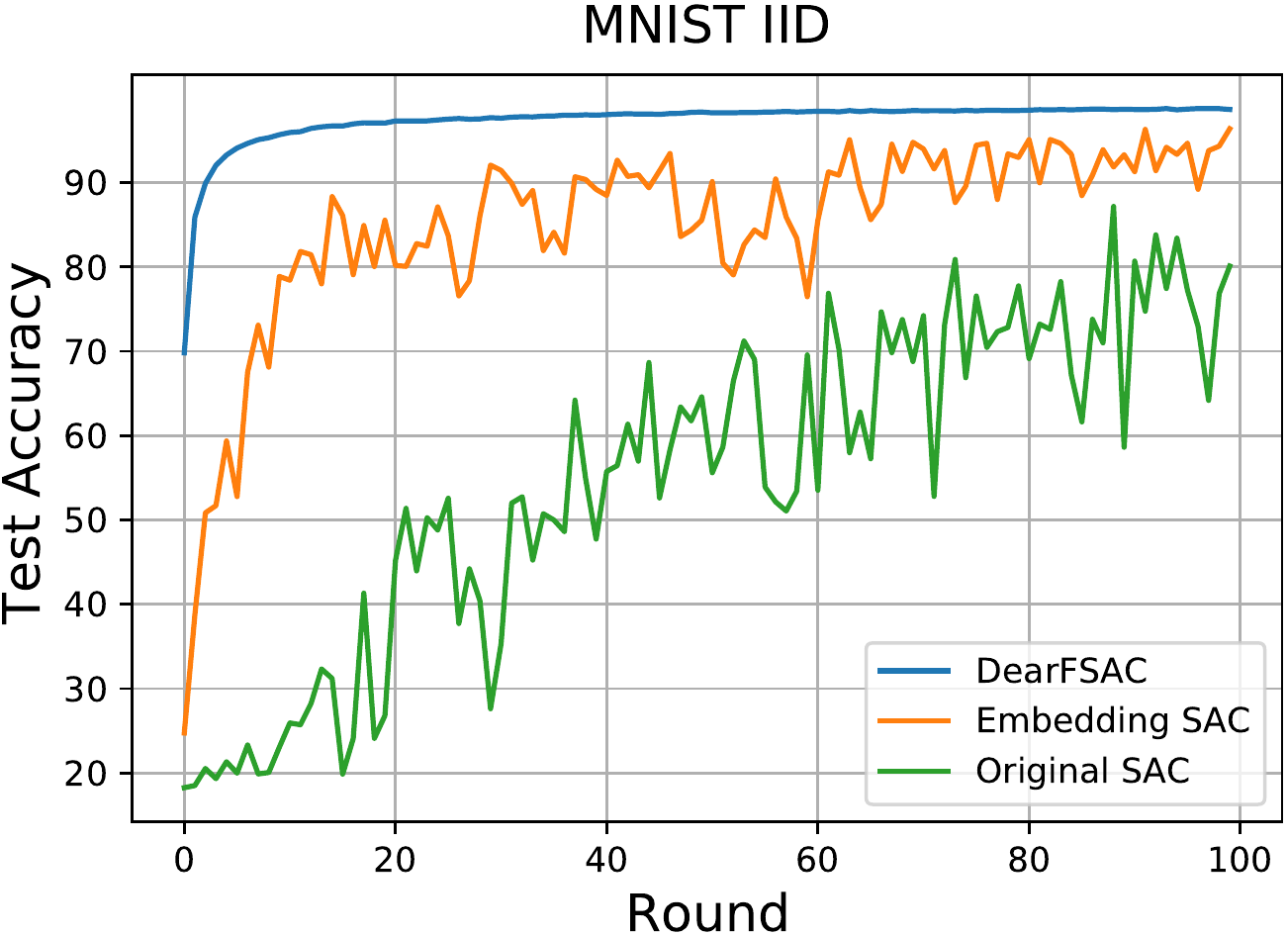}  }  
\subfloat {\includegraphics[width=0.21\textwidth]{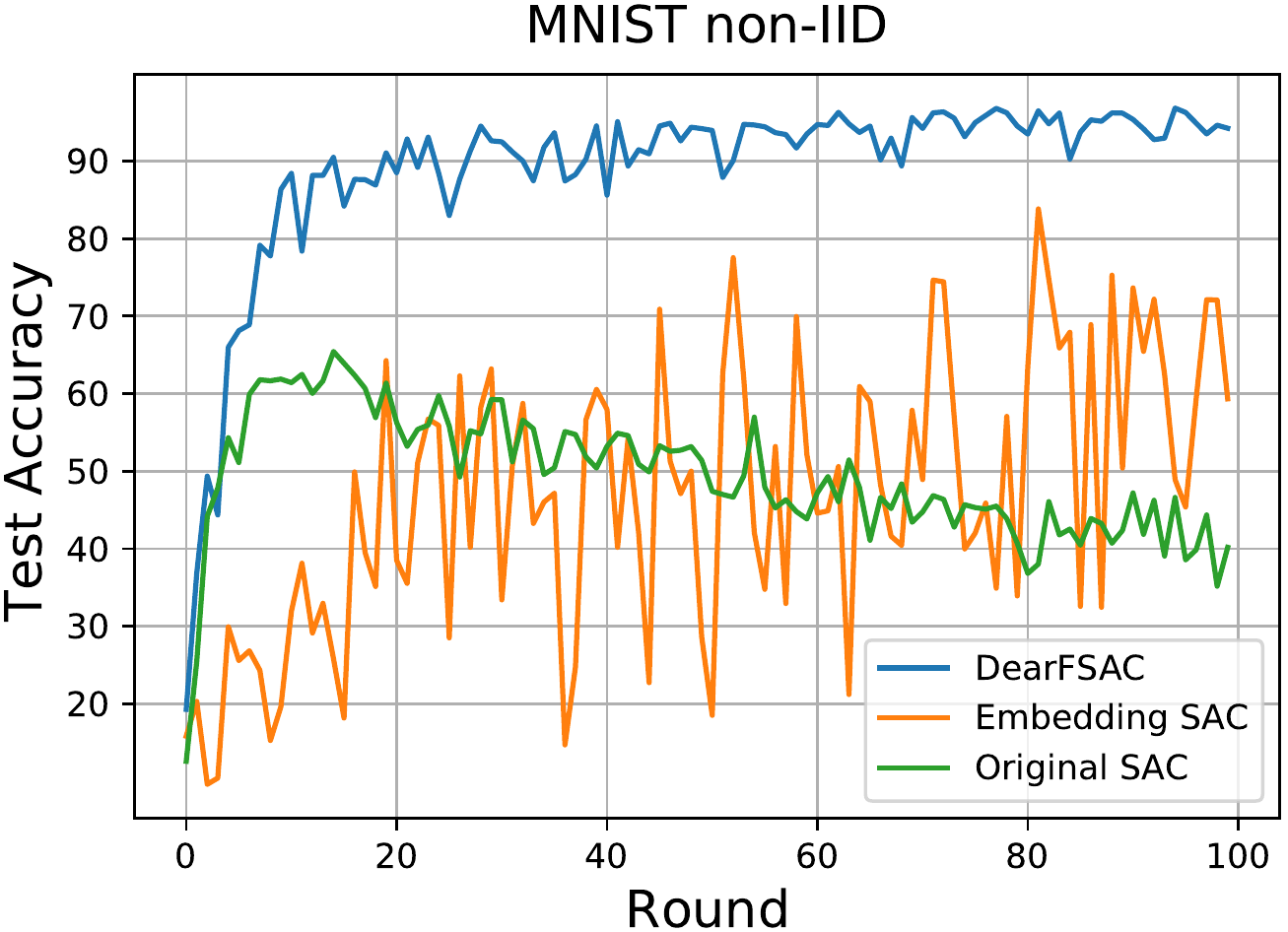}}     
\caption{The accuracy of DearFSAC, embedding SAC and original SAC on IID MNIST and non-IID MNIST, where $M=9$ and $d_{\mathcal{N}}=0.1$.}   
\label{acc_sac}     
\end{figure}

In Fig. \ref{acc_sac}, the test accuracy of DearFSAC is significantly higher than the original SAC and embedding SAC, which again proves our conclusions.

\section{Conclusion and Future Work}
In this paper, we propose DearFSAC, which assigns optimal weights to local models to alleviate performance degradation caused by defects. For model quality evaluation and dimension reduction, an auto-encoder named QEEN is designed. After receiving embedding vectors generated from QEEN, the DRL agent optimizes the assignment policy via SAC algorithm. In the experiments, we evaluate the performance of DearFSAC on four image datasets in different settings. The results show that DearFSAC outperforms FedAvg, rule-based strategy, and SL model. Specially, our model exhibits high accuracy, stable convergence, and fast training speed no matter whether there exist defects in FL process or not.

In the future, it is worthwhile investigating how to extend DearFSAC to a multi-agent framework for personalized FL in defective situations.

\bibliographystyle{named}
\bibliography{ijcai22}
\end{document}